\documentclass[11pt,a4paper]{article}
\usepackage[hyperref]{acl2020}
\usepackage{times}
\usepackage{latexsym}
\usepackage{xspace}
\usepackage{url}
\usepackage{mathtools}
\usepackage{amssymb}
\usepackage{footnote}
\usepackage{tablefootnote}
\usepackage{graphicx}
\usepackage{color, colortbl}
\usepackage{xstring}
\usepackage{comment}
\usepackage{array,booktabs,makecell}

\usepackage{microtype}

\aclfinalcopy 
\usepackage[utf8]{inputenc}
\usepackage[english]{babel}
\usepackage{fancyhdr}
\usepackage{lastpage}
 
\title{DiaNet: BERT and Hierarchical Attention Multi-Task Learning of Fine-Grained Dialect}

\author{Muhammad Abdul-Mageed$^1$  Chiyu Zhang$^1$  AbdelRahim Elmadany$^1$  Arun Rajendran$^1$  Lyle Ungar$^2$  \\ 
  $^1$Natural Language Processing Lab,  University of British Columbia \\
  $^2$Computer and Information Science, University of Pennsylvania \\
  {$^1$\tt muhammad.mageed@ubc.ca $^2$\tt  ungar@cis.upenn.edu}\\}

\begin{document}
\maketitle
\begin{abstract}
 Prediction of language varieties and dialects is an important language processing task, with a wide range of applications. For Arabic, the native tongue of $\sim 300$ million people, most varieties remain unsupported. To ease this bottleneck, we present a very large scale dataset covering 319 cities from all 21 Arab countries. We introduce a hierarchical attention multi-task learning (HA-MTL) approach for dialect identification exploiting our data at the city, state, and country levels. We also evaluate use of BERT  on the three tasks, comparing it to the MTL approach. We benchmark and release our data and models.  
\end{abstract}

\section{Introduction}
Language identification (LID) is a critical first step for multilingual NLP. Especially for processing social media such as Twitter text in global settings, the ability to identify languages, language varieties, and dialects is indispensable. In addition to classical applications of LID as an enabling technology in tasks such as machine translation, web data collection and search, and pedagogical applications ~\cite{jauhiainen2018automatic}, LID has essential real-time applications as a source of information for tracking health and well-being trends ~\citep{paul2011you}. However, of the world's currently known 7,111 living languages,~\footnote{Source: \url{https://www.ethnologue.com}.} the great majority are yet to be supported by NLP tools such as LID. As technology continues to play an increasingly impactful role in our lives, access to nuanced NLP tools (including LID) becomes an issue of equity~\cite{jurgens2017incorporating}.

\begin{figure}[h]
\center 
\frame{\includegraphics[width=8cm,height=5.5cm]{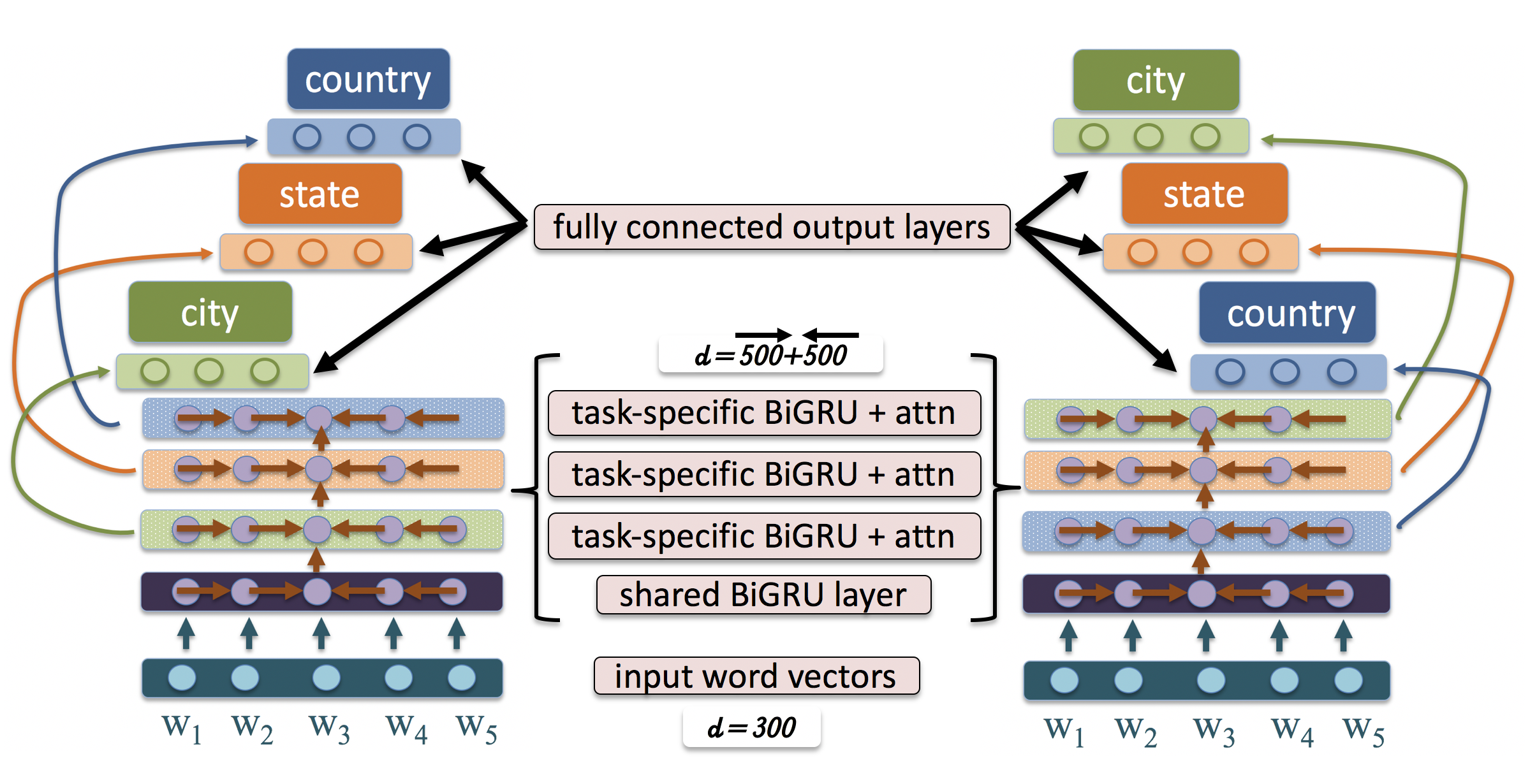}}
\label{fig:cities}
\caption{Hierarchical Attention MTL of city, state, and country. All models share one BiGRU layer of 1,000 units. Layers 2-4 are also BiGRU layers, with multi-head attention. \textbf{Left:} City network supervised at layer 2, state at layer 3, and country at layer 4. \textbf{Right:} Supervision is reversed from left network.}\label{fig:mtl2b}
\end{figure}
In spite of this key role of LID, it is still challenging to find tools for closely related languages and varieties, including those that are widely spoken. We focus on one such situation for the Arabic language, a large collection of similar varieties with $\sim$ 300 million native speakers. For Arabic, currently available NLP tools are limited to the standard variety of the language, Modern Standard Arabic (MSA), and a small set of dialects such as Egyptian, Levantine, and Iraqi. Arabic dialects differ amongst themselves and from MSA at various levels, including phonological and morphological~\cite{watson2007phonology}, lexical~\cite{salameh2018fine,mageedYouTweet2018,qwaider2018shami}, syntactic~\cite{benmamoun2011agreement}, and sociological~\cite{bassiouney2009arabic,bassiouney2017identity}. A major limitation to developing robust and equitable LID technologies for Arabic has been absence of large, diverse data. A number of pioneering efforts, including shared tasks~\citep{zampieri2014report,malmasi2016discriminating,zampieri2018language}, have been invested to bridge this gap by collecting datasets. However, these works either depend on automatic geocoding of user profiles~\citep{mageedYouTweet2018}, which is not quite accurate, as we show in Section~\ref{sec:data}, use a small set of dialectal seed words as a basis for the collection~\citep{zaghouani2018arap,qwaider2018shami}, which limits text diversity, or are based on translation of a small dataset of sentences rather than naturally-occurring text~\cite{salameh2018fine}. 

To alleviate this bottleneck, we use \textit{location as a surrogate for dialect} to build a very large scale Twitter dataset ($\sim$ 6 billion tweets), and (1) automatically label a subset of it ($\sim$  500M tweets) with coverage for all 21 Arab countries at the nuanced levels of state and city (i.e., \textit{micro-dialects}). We also (2) manually label another subset ($\sim$ 2M tweets from $\sim$ 5,000 users). We then develop highly effective supervised and weakly-supervised models exploiting the data at all three nuanced levels of city, state, and country. For modeling, we introduce a novel hierarchical attention multi-task learning (HA-MTL) network that is suited to our task (shown in Figure~\ref{fig:mtl2b}), which proves highly successful. We further investigate the newly-proposed Bidirectional Encoder Representations from Transformers (BERT) ~\citep{devlin2018bert} and show its effectiveness. 

Concretely, we make the following contributions: (1) We collect a large-scale dataset covering all Arabic varieties; (2) we introduce supervised and weakly-supervised HA-MTL models exploiting our data at fine-grained levels; (3) we empirically evaluate BERT on our tasks, showing its effectiveness; and (4) we benchmark and release our data and models. 
The rest of the paper is organized as follows: In Section~\ref{sec:data}, we introduce our Twitter data, quality assurance methods, and the external data we use for comparisons. Section~\ref{sec:methods} describes our methods. We present our supervised models in Section~\ref{sec:sup_exps} and weakly-supervised models in Section~\ref{sec:noisy}. We compare to other works in Section~\ref{sec:comparisons}, evaluate our models at the user level in Section~\ref{sec:user_eval}, review related works in Section~\ref{sec:lit}, and conclude in Section~\ref{sec:conc}.

\section{Data}\label{sec:data}
\subsection{Creating a Large User-Level Collection} To develop a large scale dataset of Arabic varieties, we extracted $\sim$ 7.5 million Twitter user ids from several in-house Arabic Twitter corpora. The corpora were collected with the Twitter streaming API, including using bounding boxes around the Arab world. The data span $\sim$ 10 years (2009-2019). We then use the Twitter API to crawl up to 3,200 tweets from a random sample of $\sim$ 2.7 million users from the collection. Overall, we acquired $\sim$ 6 billion tweets.
\begin{figure*}[h]
\center 
\frame{\includegraphics[width=10cm,height=4.5cm]{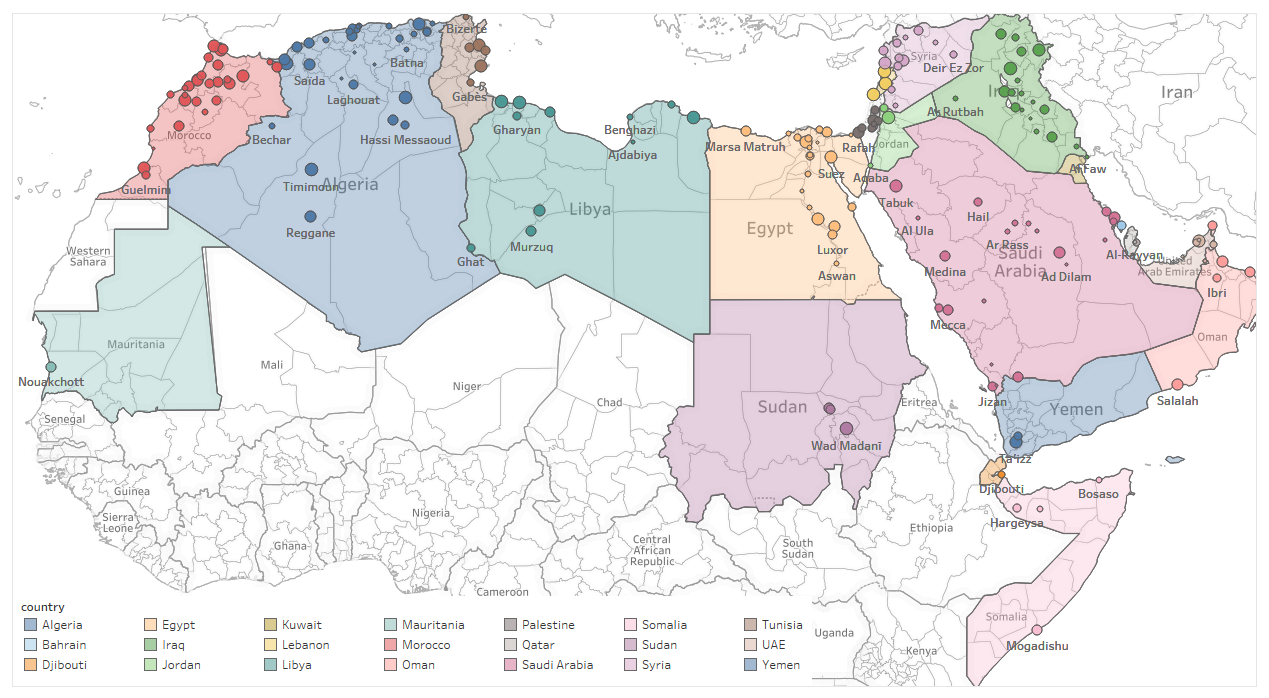}}
\caption{A map of all 21 Arab countries. States are demarcated in thin black lines within each country. A total of 319 cities (from our user location validation study, in colored circles) are overlayed within corresponding countries.}
\label{fig:all_loc}
\end{figure*}

\subsection{Automatic City Tagging} 
We use the Python geocoding library \textit{geopy} \footnote{\url{https://github.com/geopy}.} to identify the user countries (e.g., Morocco) and cities (e.g., Beirut). Geopy is a client for several popular geocoding web services aiming at locating the coordinates of addresses, cities, countries, and landmarks across the world using third-party geocoders. In particular, we use the Nominatim geocoder for OpenStreetMap data~\footnote{\url{https://nominatim.openstreetmap.org}.}. With Nominatim, Geopy depends on user-provided geographic information in Twitter profiles such as names of countries or cities to assign user location. Out of the 2.7 million users, we acquired \textit{both} `city' and `country' label for 233,105 users who contribute 507,318,355 tweets. The total number of cities initially tagged was 705, but we manually map them to only 646 as we explain next.

\subsection{Correction of City and State Tags}
\textbf{City-Level.} Investigating examples of the geolocated data, we observed geopy made some mistakes. To solve the issue, we decided to manually verify the information returned from geopy on all the 705 assumed `cities'. For this purpose of manual verification, we use Wikipedia, Google maps, and web search as sources of information while checking city names. We found that geopy made mistakes in 7 cases as a result of misspelled city names in the queries we sent (as coming from user profiles). We also found that 44 cases were not assigned the correct city name as the first `solution'. Geopy provided us with a maximum of 7 solutions for a query, with best solutions sometimes being names of hamlets, villages, etc., rather than cities. In many cases, we found the correct solution to fall between the 2nd and 4th solutions. A third problem was that some city names (as coming from user profiles) were written in non-Arabic (e.g., English or French). We solved this issue by requiring geopy to also return the English version of a city name, and \textit{exclusively} using that English version. Ultimately, we acquired a total of 646 cities.

\textbf{State-Level.} 
Geopy also returned to us a total of 192 states/provinces that correspond to the 646 cities. We manually verified all the state names, and their correspondence to the cities and countries and found no issues. 

\subsection{Data Pre-processing} 
To keep only high-quality data, we apply the following procedures: First, we remove all re-tweets, decreasing the collection to 318,174,122 tweets. Second, we normalize the tweets by reducing 2 or more consecutive sequences of the same character to only 2, replace usernames with $<USER>$ and URLs with $<URL>$. Finally, we remove all tweets with less than three actual \textit{Arabic} words. This further reduces the collection to 277,430,807 tweets. Since for most Arabic varieties there are no available tokenizers, we tokenize input text only lightly by splitting off punctuation.
\subsection{Validation of User Location}
After manually correcting the city and state names, we needed to verify that a given user actually belongs to the automatically assigned location labels (city, state, and country). To achieve this, we first excluded cities that have $< 500$  tweets and users with $<30$ tweets from the data. This gave us 319 cities. We then ask two native Arabic annotators to label the data. Their job was to consider the automatic label for each task (city and country)~\footnote{Note that we have already manually established the link between states and their corresponding cities and countries.} and assign one label from the set \textit{\{true, false, unknown\}} per task for each user in the collection. We trained the annotators and instructed them to examine the profile information of each user on Twitter, providing a link to the profile. We asked them to consider various sources of information as a basis for their decisions, including (1) the profile picture, (2) profile textual description (including user-provided location), (3) the actual name of the user (if available), (4) at least 10 tweets, (5) the followers and followees of the user, and (5) user's network behavior such as the `likes'. Each annotator was responsible for $\sim 50\%$ of the usernames and was given a random sample of 20 users for each city along with the Twitter handles and the automatically assigned \textit{city} and \textit{country} labels. We asked the users to label the first 10 accounts in each city, and only add more if the city proves specially challenging (as we observed to be the case in a pilot analysis of a few cities). Annotators ended up labeling a total of 4,953 accounts, of whom 4,012 users were verified for \textit{both} country and city locations. We found that 81.00\% of geopy tags for country are correct, but only 62.29\% for city. As a final sanity check, a third annotator reviewed the labels for a random sample of 20 users from each annotator and agreed fully. Figure~\ref{fig:all_loc} shows a map of all 21 Arab countries, each divided into its states with cities overlayed as colored small circles. We now describe the external datasets we use for comparisons.

\begin{table}[]
\centering
\footnotesize 
\begin{tabular}{lccl}
\hline
\textbf{Country} & \textbf{\%vld\_cntry} & \textbf{\%vld\_city} & \textbf{\#tweets} \\ \hline
\textbf{Algeria} & 77.49\ & 69.74\ & 185,854 \\
\textbf{Bahrain} & 83.95\ & 39.51\ & 25,495 \\
\textbf{Djibouti} & 68.42\ & 68.42\ & 3,939 \\
\textbf{Egypt} & 92.66\ & 64.02\ & 463,695 \\
\textbf{\dots} & \dots\ & \dots\ & \dots \\
\textbf{Yemen} & 72.41\ & 56.32\ & 47,450 \\
\textbf{Avg/Total} & 81.00\ & 62.29\ & 2,025,013 \\\hline
\end{tabular}
\caption{A subset of our gold data from manually verified users.}
\label{tab:my-table}
\end{table}

\subsection{External Data}\label{subsec:external_data}

\textbf{Arap-Tweet} ~\cite{zaghouani2018arap} comprises 17 countries collected from 1,100 manually-verified Twitter users based on a seed-word approach. The dataset totals 2.4M tweets. In comparison, our dataset covers more countries, has more nuanced tags (on cities and states), and is extracted from more users, thus making it more diverse (since we also do not use seed words to find our users). ~\citet{zaghouani2018arap} do not perform classification exploiting their data. We split Arap-Tweet into 80\% TRAIN, 10\% DEV, and 10\% TEST.
\textbf{SHAMI} ~\cite{qwaider2018shami} is a Twitter and web fora dataset of Jordanian, Lebanese, Palestinian, and Syrian Arabic collected with a seed-word approach. It has 66,249 manually labeled tweets. In comparison, our dataset is much larger, covers more countries, and is more diverse. 
We split SHAMI into TRAIN (80\%), DEV (10\%), and TEST (10\%) for our experiments, thus using less training data than ~\citet{qwaider2018shami} who employ cross-validation.

\textbf{MADAR Shared Task-2} ~\cite{bouamor2019madar_sharedtask} is a dataset released for the MADAR Twitter User Dialect Identification Shared Task 2. The dataset is distributed as train, dev, and test (without labels) with user and tweet ids. We were able to crawl the data for a total of 2,311 users, acquiring $193,086$,$26,588$, and $43,909$ tweets for the three splits, respectively. We call training data TRAIN-I as we also create another training set (TRAIN-II) that is a concatenation of task 2 and task 1 data.~\footnote{Task 1 is also organized by~\citet{bouamor2019madar_sharedtask}.}
\section{Methods}\label{sec:methods}
We perform dialect identification at the country, state, and city levels. We use two main classification methods, Gated Recurrent Units (GRUs)~\cite{cho2014learning}, a variation of recurrent neural networks (RNN), and Google's bidirectional masked language model based on transformers (BERT)~\cite{devlin2018bert}. We now describe each of these methods.

\subsection{GRU}
A Gated Recurrent Unit (GRU)~\cite{cho2014learning} is a type of cell proposed to simplify recurrent neural network (RNN) learning. It makes use of an \textit{update gate} $\textbf{\textit{z}}^{(t)}$ and a \textit{reset gate} $\textbf{\textit{r}}^{(t)}$. The activation of GRU at time step $t$ is a linear interpolation of the previous activation \textit{hidden state} $\textbf{\textit{h}}^{(t-1)}$ and the candidate activation \textit{hidden state} \begin{math}  \textbf{\textit{$\widetilde{h}$}}^{(t)} \end{math}. The \textit{update state} $\textbf{\textit{z}}^{(t)}$ decides how much the unit updates its content, and the candidate activation makes use of a \textit{reset gate} $\textbf{\textit{r}}^{(t)}$. When its value is close to zero, the reset gate allows the unit to \textit{forget} the previously computed state.
\subsection{Multi-Task Learning}

\begin{figure}[h]
\centering
\frame{\includegraphics[width=0.4\textwidth]{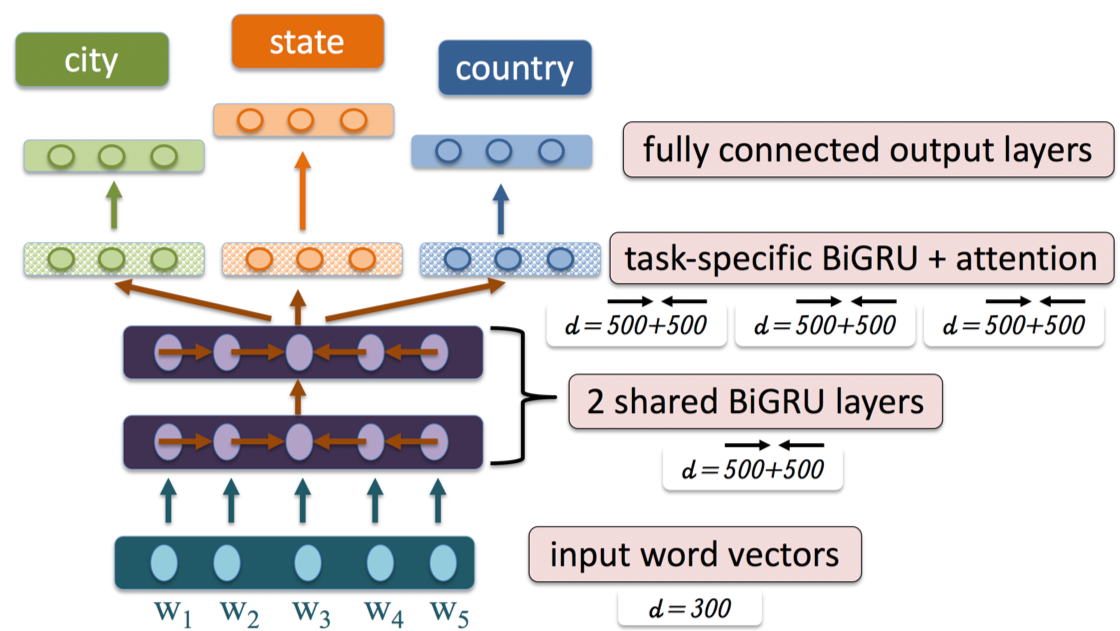}}
\caption{Our MTL network for city, state, and country. The three tasks share 2 hidden layers, with each task having its independent attention layer.}
\label{fig:mtl_t_spec_attn}
\end{figure}
We investigate the utility of \textit{multi-task learning} (MTL) for language ID. The intuition behind MTL is that many real-world tasks involve making predictions about closely related labels or outcomes. For related tasks, MTL helps achieve inductive transfer between the various tasks by leveraging additional sources of information from some of the tasks to improve performance on the target task~\cite{caruana1993}. By using training signals for related tasks, MTL allows a learner to prefer hypotheses that explain more than one task~\cite{caruana1997multitask} and also helps regularize models.

In single task learning, an independent network is trained in isolation for each task.
In contrast, in MTL, a number of tasks are learned together in a single network, with each task having its own output. An MTL network has a shared input, and one or more hidden layers that are shared between all the tasks. Backpropagation is then applied in parallel on all outputs. In our case, we train a single network for our city, state, and country tasks with one output for each of the three tasks. Figure ~\ref{fig:mtl_t_spec_attn} is an illustration of an MTL network for our 3 tasks, with 2 shared hidden BiGRU layers and a task-specific (i.e., independent) BiGRU attention layer. In our current work, each of the three tasks has its own loss function, with the MTL loss computed as:

\begin{equation}
\begin{multlined}
 \mathcal{L}(\theta_{MTL})\\= \left(\mathcal{L}(\theta_{city}) + \mathcal{L}(\theta_{state})+\mathcal{L}(\theta_{country}) \right) / 3 \\
 \end{multlined}
\end{equation}

We now introduce the Transformer~\cite{vaswani2017attention}, since both our attention mechanism and BERT ~\cite{devlin2018bert} are based on it.

\subsection{Transformer}\label{subsec:transformer}
The Transformer ~\citep{vaswani2017attention} is based solely on attention. Similar to most other sequence transduction models~\cite{bahdanau2014neural,cho2014learning,Sutskever2014SequenceTS}, it is an encoder-decoder architecture. It takes a sequence of symbol representations $x^{(i)} \dots x^{(n)}$, maps them into a sequence of continuous representations $z^{(i)} \dots x^{(n)}$ that are then used by the decoder to generate an output sequence $y^{(i)} \dots y^{(n)}$, one symbol at a time. This is performed using \textit{self-attention}, where different positions of a single sequence are related to one another. The Transformer employs an attention mechanism based on a function that operates on \textit{queries}, \textit{keys}, and \textit{values}. The attention function maps a query and a set of key-value pairs to an output, where the output is a weighted sum of the values. For each value, a weight is computed as a compatibility function of the query with the corresponding key. We implement and apply the multi-head attention function to our BiGRU models. 

\textit{Encoder} of the Transformer in~\citet{vaswani2017attention} has 6 attention layers, each of which is composed of two sub-layers: (1) \textit{multi-head attention} where, rather than performing a single attention function with queries, keys, and values, these are projected \textit{h} times into linear, learned projections and ultimately concatenated; and 
(2) fully-connected \textit{feed-forward network (FFN)} that is applied to each position separately and identically. 
\textit{Decoder} of the Transformer also employs 6 identical layers, similar to the encoder, yet with an extra/third sub-layer that performs multi-head attention over the encoder stack. 
As mentioned, the Transformer is the core learning component in BERT~\cite{devlin2018bert}, which we now introduce.

\subsection{BERT}\label{subsec:bert}
BERT~\cite{devlin2018bert} stands for \textbf{B}idirectional \textbf{E}ncoder \textbf{R}epresentations from \textbf{T}ransformers. It is an approach for pre-training language representations that involves two unsupervised learning tasks, (1) \textit{masked language models (Masked LM)} and (2) \textit{next sentence prediction}. Since BERT uses bidirectional conditioning, a given percentage of random input tokens are masked and the model attempts to predict these masked tokens. 
.~\citet{devlin2018bert} mask 15\% of the tokens (the authors use \textit{word pieces}) and feed the final hidden vectors of these masked tokens to an output softmax over the vocabulary. The next sentence prediction task of BERT is also straightforward.~\citet{devlin2018bert} simply cast the task as binary classification. For a given sentence \texttt{S}, two sentences \texttt{A} and \texttt{B} are generated where \texttt{A} (positive class) is an actual sentence from the corpus and \texttt{B} is a randomly chosen sentence (negative class). Once trained on an unlabeled dataset, BERT can then be fine-tuned with supervised data.

\section{Gold-Supervised Models}\label{sec:sup_exps}

\begin{table*}[h]
\centering
\begin{tabular}{|l|ll|ll|ll|}
\hline
\textbf{Setting} & \multicolumn{2}{c|}{\textbf{City}} & \multicolumn{2}{c|}{\textbf{State}} & \multicolumn{2}{c|}{\textbf{Country}} \\ \hline
\textbf{Eval Metric} & \textbf{acc} & \textbf{F1} & \textbf{acc} & \textbf{F1} & \textbf{acc} & \textbf{F1} \\ \hline
\textbf{Baseline I (majority in TRAIN)} & 1.313 & 0.008 & 3.110 & 0.032 & 9.191 & 0.802 \\
\textbf{Baseline II (single task Attn-BiGRU)} & 2.740 & 0.880 & 4.450 & 0.910 & 27.170 & 12.820 \\ \hline
\textbf{MTL (common-attn)} & 4.036 & 1.693 & 5.693 & 2.195 & 28.255 & 13.362 \\
\textbf{MTL (spec-attn)} & 4.000 & 1.479 & 5.956 & 2.085 & 28.946 & 13.858 \\
\textbf{HA-MTL (city first)} & \underline{12.295} & \underline{11.736} & \underline{13.728} & 12.836 & 40.349 & \underline{29.869} \\
\textbf{HA-MTL (country first)} & 11.265 & 10.588 & 13.577 & \underline{12.869} & \underline{41.250} & 29.763 \\ \hline
\textbf{BERT} & \textbf{19.329} & \textbf{19.452} & \textbf{19.329} & \textbf{19.452} & \textbf{47.743} & \textbf{38.122}\\ \hline
\end{tabular}%
\caption{Performance on TEST. Highest results for MTL are \underline{underlined}. BERT results (best) are in \textbf{bold}.}
\label{tab:sprvsd_eval}
\label{tab:my-table}
\end{table*}
In this section, we explain how we split our data and specify our baseline and evaluation metrics. We then present our gold-supervised models, namely (i) our single-task models (Section ~\ref{subsec:stl}), (ii) multi-task models (Section ~\ref{subsec:mtl}) and (iii) our BERT model (Section ~\ref{subsec:bert}). 

\subsection{Data Splits, Baseline, and Evaluation}\label{subsec:splits}
We randomly split our own manually-verified dataset into 80\% training (TRAIN), 10\% development (DEV), and 10\% test (TEST). To limit the GPU hours needed for processing, we cap the number of tweets in our TRAIN in any given country at 100K. This reduces the TRAIN size from 1,620,436 to 1,099,711. Our DEV set has 202,509 tweets, and our TEST has 202,068 tweets.~\footnote{The distribution of classes in our splits is in the supplementary material.} 
For all our experiments, we remove diacritics from the input text. We use \textit{two baselines}: the majority class in TRAIN (Baseline I) and a single-task BiGRU (Baseline II, described in Section~\ref{subsec:stl}).
For all our experiments, we tune model hyper-parameters and identify best architectures on DEV. We run all models for 15 epochs, with early stopping `patience' value of 5 epochs, choosing the model that performs highest on DEV as our best model. We then run each best model on TEST, and report \textit{accuracy} and \textit{macro $F_{1}$ score}.~\footnote{We include a table with results on DEV in the supplementary material.}

\subsection{Single-Task BiGRUs}\label{subsec:stl}

As a \textit{second baseline} (Baseline II), we build an independent network for each of the 3 tasks using the same architecture and model capacity. Each network has 3 hidden BiGRU layers,~\footnote{We also ran single-task networks with 4 hidden layers, but we find them to overfit quickly even when we regularize with dropout at $0.7$ on all layers.} with 1,000 units each (500 units from left to right and 500 units from right to left). We add multi-head attention \textit{only} to the third hidden layer. We trim each sequence at 50 words,~\footnote{In initial experiments, we found a maximum sequence of $30$ words to perform slightly worse.} and use a batch size of 8. Each word in the input sequence is represented as a vector of 300 dimensions that are learned directly from the data. Word vectors weights $W$ are initialized with a standard normal distribution, with $\mu=0$, and $\sigma=1$, i.e., $W \sim N(0,1)$. 
For optimization, we use Adam~\cite{kingma2014adam} with a fixed learning rate of $1e-3$. For regularization, we use dropout~\cite{srivastava2014dropout} with a value of $0.5$ on each of the 3 hidden layers. Table~\ref{tab:sprvsd_eval} presents our results on TEST.

\subsection{MTL}\label{subsec:mtl}
With MTL, we design a single network to learn the 3 tasks simultaneously. In addition to our hierarchical attention MTL (HA-MTL) network, we design two architectures that differ as to how we endow the network with the \textit{attention mechanism}. We describe these next.

\subsubsection{Shared and Task-Specific Attention}
We first design networks with attention at the same level in the architecture. Note that we use the same hyper-parameters as the single-task networks. We have two configurations:

\textbf{Shared Attention:}
In this configuration, we design a network with 3 hidden BiGRU layers, each of which has 1,000 units per layer (500 in each direction).~\footnote{Again, 4 hidden-layered network for both the \textit{shared} and \textit{task-specific} attention settings were sub-optimal and so we do not report their results here.} All the 3 layers are shared across the 3 tasks, including the third layer. Only the third layer has attention applied. We refer to this setting as \textit{MTL-common-attn}. 

\textbf{Task-Specific Attention:}
This network is similar to the previous one in that the first two hidden layers are shared, but differs in that the third layer (attention layer) is task-specific (i.e., independent for each task). We refer to this setting as \textit{MTL-spec-attn}. Figure~\ref{fig:mtl_t_spec_attn} illustrates our MTL network for learning city, with task-specific attention. This architecture allows each task to specialize its own attention within the same network. As Table~\ref{tab:sprvsd_eval} shows, both \textit{MTL-common-attn} and \textit{MTL-spec-attn} improve over each of the two baselines, and are consistently complimentary: While the first acquires better acc, the latter is slightly better in $F_1$ score. 

\subsection{Hierarchical Attention MTL (HA-MTL)}

We design a single network for the 3 tasks, but with supervision at different layers. Overall, the network has 4 BiGRU layers (each with a total of 1,000 units), the bottom-most of which has no attention. Layers 2, 3, and 4 each has multi-head attention applied, followed directly by one task-specific fully-connected layer with softmax for class prediction. This is the architecture depicted in Figure~\ref{fig:mtl2b}. On the left side of Figure~\ref{fig:mtl2b}, we show the \textit{city-first} hierarchical attention network, with city supervised at the second hidden layer. On the right side, we have the \textit{country-first} network, where country is supervised earlier (at the second layer). In the two scenarios, state is supervised at the middle layer. These two architectures allow information flow with different granularity: While the city-first network tries to capture what is in the physical world a more fine-grained level (city), the country-first network does the opposite. Again, we use the same hyper-parameters as the single-task and MTL networks, but we use a dropout rate of 0.70 since we find it to work better. As Table~\ref{tab:sprvsd_eval} shows, our proposed HA-MTL models significantly outperform single-task and other MTL models. They outperform our Baseline II with 9.555\%, 9.277\%, and 14.079\% acc on city, state, and country prediction respectively, thus demonstrating their effectiveness on the task.

\subsection{BERT Models}\label{subsec:bert_models}
We use the BERT-Base, Multilingual Cased model released by the authors~\footnote{\url{https://github.com/google-research/bert/blob/master/multilingual.md}.}. The model is trained on 104 languages, including Arabic, with 12 layer, 768 hidden units each, 12 attention heads, and has 110M parameters. The model has 119,547 word pieces for each language. For fine-tuning, we use a maximum sequence size of 50 words and a batch size of 32. We set the learning rate to 2e-5. We train for 15 epochs, as mentioned earlier. As Table~\ref{tab:sprvsd_eval} shows, BERT performs consistently better on the three tasks. It outperforms the best of our two HA-MTL networks with an acc of 7.034\% (city), 5.601\% (state), 6.493\% (country). And $F-_{1}$ of 7.716, 6.583, and 8.253 for the 3 tasks, respectively.

\section{Learning From Noisy Labels}\label{sec:noisy}

\begin{table}[h]
\centering
\footnotesize 
\begin{tabular}{lll} \hline 
\textbf{Supervision} & \textbf{acc} & \textbf{F1} \\ \hline 
\textbf{Baseline I (majority in TRAIN) } & 9.207 & 0.843 \\
\textbf{Baseline II (Gold)} & 46.844 & 37.643 \\ \hline
\textbf{Weak} & 41.166 & 23.697 \\
\textbf{Weak+Gold} & \textbf{49.768} & 38.254 \\
\textbf{Weak\_\textit{then}\_Gold} & 47.862 & \textbf{38.560}\\ \hline
\end{tabular}%
\caption{Results on TEST with models exploiting noisy labels on 20 countries (with Djibouti excluded). For comparison, our gold (BERT trained on human-labeled TRAIN) is re-trained with 20 classes.} 
\label{tab:noisy_res}
\end{table}
In contrast to our gold-supervised models (Section~\ref{sec:sup_exps}), this set of experiments is focused on learning from noisy labels. We only perform experiments on predicting \textit{country} labels. Our goal is to answer the question ``To what extent can automatically acquired labels in our dataset be beneficial for learning?". To this end, we remove human annotated users from our larger automatically labeled pool and use only data tagged with any of the 319 cities whose users we manually labeled. Keeping tweets with at least 3 Arabic words, we acquire 1,161,651 tweets from 3,195 users, across all countries except Djibouti (all whose users were already in our human annotation round). As such, we have 20 countries in this dataset and refer to it simply as \texttt{Auto-Tagged}. We exploit Auto-Tagged in 3 experimental settings, reporting results on our gold TEST in all 3 cases. The 3 settings are: \textbf{(1) Weakly Supervised:} Where fine-tune BERT exclusively on Auto-Tagged; (2) \textbf{Weak+Gold:} Where concatenate Auto-Tagged and our TRAIN (gold), shffle the dataset, and fine-tune BERT on it; and (3) \textbf{Weak-\textit{Then}-Gold:} Where fine-tune BERT on Auto-Tagged \textit{first}, then resume fine-tuning on our human labeled data (TRAIN). Table~\ref{tab:noisy_res} shows \textbf{Weak+Gold} to improve 2.923\% acc over our \textbf{Gold} model (Baseline II), establishing the utility of using noisy labels on the country level.  
\section{Comparisons to Other Models}\label{sec:comparisons}
Since the existing data described in Section~\ref{subsec:external_data} have varying numbers of classes (different from our data), we train BERT on their respective TRAIN splits (as described in Section~\ref{subsec:external_data}). While ~\citet{qwaider2018shami} use linear classifiers to model their data, there are no models we know of for Arap-Tweet~\cite{zaghouani2018arap} nor MADAR~\cite{bouamor2019madar_sharedtask}. As such, we publish the first results on these two datasets. As a baseline, we run a unidirectional 1-layered GRU, with 500 units, on each of Arap-Tweet and MADAR.~\footnote{We evaluate on MADAR DEV set.}

As Table~\ref{tab:external_eval} shows, our models outperform ~\citet{qwaider2018shami} on SHAMI.
We also establish new results on both Arap-Tweet and MADAR. Note that we do not report on the dataset described in ~\citet{mageedYouTweet2018} since it is automatically labeled, and so is \textit{noisy}. We also do not compare to the dataset in ~\citet{salameh2018fine} since it is small and \textit{not naturally occurring} (2,000 translated sentences per class).~\footnote{~\citet{salameh2018fine} report that linear classifiers outperform deep learning models due to small data set size.}
\begin{table}[h]
\centering
\footnotesize 
\begin{tabular}{lllll} \hline 
\textbf{Dataset} & \textbf{Model} & \textbf{\#cls} & \textbf{acc} & \textbf{F1} \\ \hline 
\textbf{ARAB-TWT} & GRU-500 & 17 & 38.787 & 39.171 \\
\textbf{} & Ours & 17 & \textbf{54.606} & \textbf{55.066} \\ \hline 
\textbf{MADAR} & GRU-500 & 21 & 46.810 & 29.840\\

\textbf{} & Ours, TRAIN-I & 21 & 48.499 & 33.929\\
\textbf{} & Ours, TRAIN-II & 21 & \textbf{49.394} & \textbf{35.931}\\\hline
\textbf{SHAMI} & Qwaider et al.18 & 4 & 70.000 & 71.000 \\
\textbf{} & Ours & 4 & \textbf{86.065} & \textbf{85.464} \\ \hline
\end{tabular}
\caption{Results on external data. Best performance and new results where there are no models to compare to are \textbf{bolded}.} 
\label{tab:external_eval}
\end{table}
\section{User-Level Evaluation}\label{sec:user_eval}
Our models are not designed to directly detect the dialect of a user, but rather takes a single tweet input at a time. However, we test how the model will fare on detecting user-level dialect given a certain number of tweets from a random user. For the purpose, we crawl up to 500 tweets from each of 500 users from the MADAR~\cite{bouamor2019madar_sharedtask} user base and extract the following number of tweets from each user: \textit{\{10, 25, 50, 75, 100, 500\}}. We run our best performing BERT model (from the 21 countries in Table~\ref{tab:sprvsd_eval}) on these user tweets, one tweet at a time. Taking the majority class on each user's tweets, we find that with 100 tweets, for example, the model can reach 65.171\% acc and with 500 tweets, it can reach 66.787\% acc.~\footnote{We provide the full results table for user-level evaluation in supplementary material.}~\footnote{In our 233,105 automatically tagged users, 94.85\% have $>=$100 tweets, suggesting a model based on only 100 tweets would have very high coverage.}
\section{Related Work}\label{sec:lit}
\textbf{Arabic Dialects.} Most of the early categorizations of Arabic dialects arbitrarily depended on cross-country geographical divisions~\cite{habash2010introduction,versteegh2014arabic}. More recent treatments such as~\citet{mageedYouTweet2018},~\citet{salameh2018fine},~\citet{qwaider2018shami},and ~\citet{zaghouani2018arap} focus on more fine-grained levels of dialectness, e.g., country and city levels. These works are more aligned with sociolinguistic work, e.g., ~\citet{labove1964} and ~\citet{trudgill1974linguistic}, showing language can vary at smaller regions such as different parts of the same city, thus creating micro-dialects within the same dialect. The finest Arabic variations treated in the literature cover 25 to 29 cities ~\cite{salameh2018fine,mageedYouTweet2018}. To the best of our knowledge, our work constitutes the most fine-grained attempt to classify Arabic varieties, including \textit{micro-dialects}. We also use a much larger dataset than previous works.

\textbf{Dialectal Arabic Data and Models.} Once primarily spoken, Arabic varieties came into written form with the proliferation of social media. Much of the early work focused on collecting data for main varieties such as Egyptian and Levantine \cite{diab2010colaba,elfardy2012simplified,al2012yadac,sadat2014automatic,zaidan2011arabic}. Many works developed models for detecting 2-3 dialects \cite{elfardy2013sentence,zaidan2011arabic,zaidan2014arabic,cotterell2014multi}. These works, e.g., ~\citet{elfardy2013sentence} and ~\citet{tillmann2014improved}, mostly exploit AOC \cite{zaidan2011arabic}. 
Larger datasets, mainly based on Twitter, were recently introduced~\cite{mubarak2014using,mageedYouTweet2018,zaghouani2018arap}. Our dataset (labeled and unlabeled) is orders of magnitude than available datasets, and by far the most fine-grained. 

\textbf{Geolocation.} Relevant to our work is also research on geolocation ~\cite{han2016twitter,do2018twitter}. Rather than predicting geolocation, we focus on urban locations such as cities and states as surrogates for micro-dialects.

\textbf{MTL.} MTL has been successfully applied to many NLP problems, including MT and syntactic parsing~\cite{luong2015multi}, sequence labeling~\cite{sogaard2016deep,rei2017semi}, and text classification~\cite{liu2016recurrent}. As we have shown, MTL is well-suited to fine-grained dialect prediction and, to the best to our knowledge, we are the first to apply it to this problem.
\section{Conclusion}\label{sec:conc}
We proposed an approach for using location as a surrogate for dialect aiming at building a very large scale Twitter dataset of Arabic varieties. Our data and models cover varieties from all 21 Arab countries, including the nuanced levels of city and state. We also introduced an effective hierarchical attention multi-task learning (HA-MTL) approach for modeling varieties and micro-dialects. Furthermore, we empirically demonstrated the utility of BERT on our tasks. In addition, we benchmarked our data and models for release and reported new state-of-the-art results on a number of external datasets. Ultimately, our work has the potential to open up opportunities for investigating variants of Arabic that remain largely understudied. The work is also a first step toward deployment of Arabic NLP technologies in real-world applications, such as in disaster and emergency situations where diverse varieties are in actual use.

\bibliography{acl2020.bib}
\bibliographystyle{acl_natbib}

\newpage
\appendix
%
%

\section{Summary of Supplementart Material}\label{sec:summary}
We provide the following supplementary items:
\begin{enumerate}
    \item Figure~\ref{fig:all_loc} is a bigger version of the map of all 21 Arab countries, with corresponding states and cities provided in the manuscript.
    \item Table~\ref{tab:data_stats} shows statistics of the dataset for 233,105 users for which we acquired geotags. 
    \item Table~\ref{tab:gold_data} provides statistics across the 21 countries of our gold data from the manually verified users.
    \item Table~\ref{tab:data_stats} provides statistics across the TRAIN, DEV, and TEST splits in our gold dataset, after capping dominant classes at 100,000 tweets each.
    \item Tables~\ref{tab:sprvsd_eval_dev} and~\ref{tab:sprvsd_eval_test} show results of our supervised models, in both DEV and TEST data. We re-produce TEST results here for convenience.
    \item Tables~\ref{tab:noisy_res_dev} and~\ref{tab:noisy_res_test} show our results from experiments exploiting noisy labels. Again, we re-produce TEST results here for convenience.
    \item Table~\ref{tab:user_eval} shows results on user-level evaluation, with different sizes of tweets per user.
\end{enumerate}
\begin{figure*}[h]
\center 
\frame{\includegraphics[width=16cm,height=8.5cm]{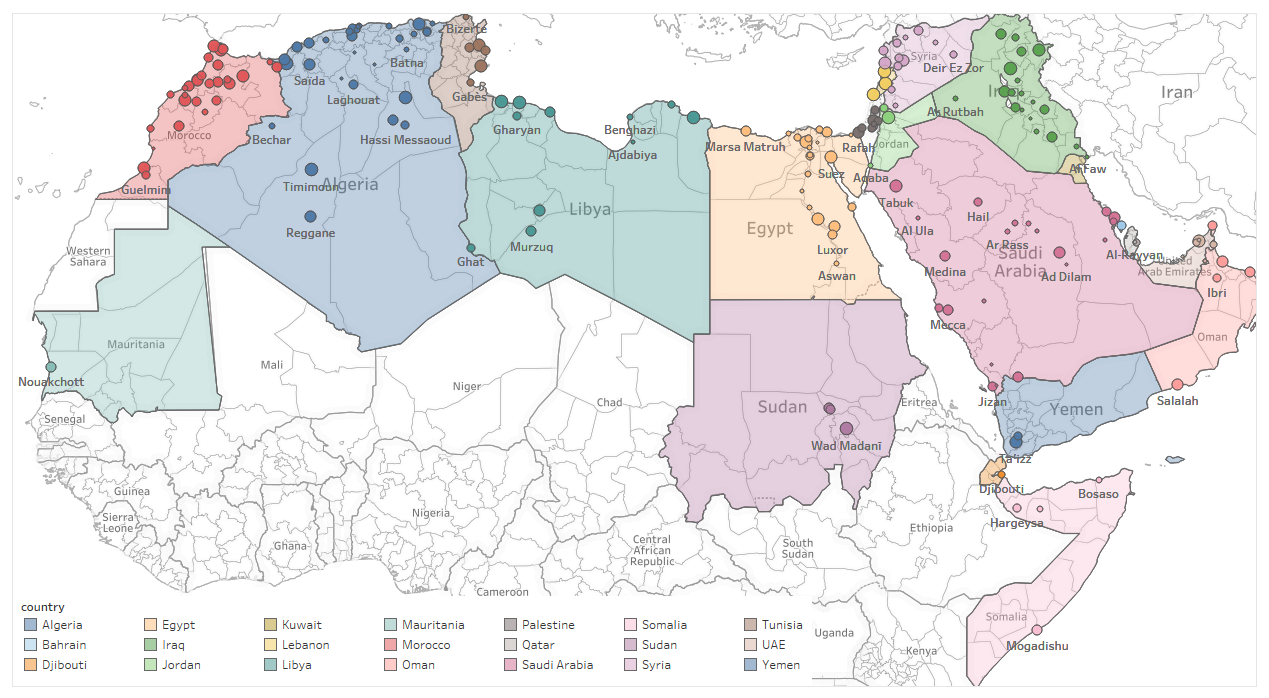}}
\caption{A bigger-sized map of all 21 Arab countries. States are demarcated in thin black lines within each country. A total of 319 cities (from our user location validation study, in colored circles) are overlayed within corresponding countries.}
\label{fig:all_loc}
\end{figure*}
\begin{table*}[t!]
\centering
\begin{tabular}{|
>{}l |
>{}c |l|l|l|l|c|c|}
\hline
\multicolumn{2}{|c|}{\textbf{Countries}} &                                    & \multicolumn{3}{c|}{\textbf{\#Tweets}}                                                                                  &                                     &                                    \\ \cline{1-2} \cline{4-6}
\textbf{Name}                   & \textbf{Code}                  & {\textbf{\#Users}} & \textbf{Collected}   & \textbf{-Retweets} & \textbf{Normalized}  & {\textbf{\#States}} & {\textbf{\#Cities}} \\ \hline
Algeria                         & dz                             & 1,960                                                      & 3,939,411                                    & 2,889,447                                         & 2,324,099                                    & 47                                                          & 200                                                         \\ 
Bahrain                         & bh                             & 1,080                                                      & 2,801,399                                    & 1,681,337                                         & 1,385,533                                    & 4                                                           & 4                                                           \\ 
Djibouti                        & dj                             & 6                                                          & 11,901                                       & 9,173                                             & 8,790                                        & 1                                                           & 1                                                           \\ 
Egypt                           & eg                             & 42,858                                                     & 92,804,863                                   & 61,264,656                                        & 47,463,301                                   & 27                                                          & 56                                                          \\ 
Iraq                            & iq                             & 4,624                                                      & 7,514,750                                    & 4,922,553                                         & 4,318,523                                    & 18                                                          & 62                                                          \\ 
Jordan                          & jo                             & 3,806                                                      & 7,796,794                                    & 5,416,413                                         & 4,209,815                                    & 4                                                           & 5                                                           \\ 
KSA                             & sa                             & 136,455                                                    & 297,264,647                                  & 177,751,985                                       & 165,036,420                                  & 11                                                          & 31                                                          \\ 
Kuwait                          & kw                             & 4,466                                                      & 11,461,531                                   & 7,984,758                                         & 6,628,689                                    & 4                                                           & 14                                                          \\ 
Lebanon                         & lb                             & 1,364                                                      & 3,036,432                                    & 1,893,089                                         & 1,160,167                                    & 6                                                           & 19                                                          \\ 
Libya                           & ly                             & 2,083                                                      & 4,227,802                                    & 3,109,355                                         & 2,655,180                                    & 21                                                          & 32                                                          \\ 
Mauritania                      & mr                             & 102                                                        & 209,131                                      & 148,261                                           & 129,919                                      & 4                                                           & 4                                                           \\ 
Morocco                         & ma                             & 1,729                                                      & 3,407,741                                    & 2,644,733                                         & 1,815,947                                    & 17                                                          & 117                                                         \\ 
Oman                            & om                             & 4,260                                                      & 8,139,374                                    & 4,866,813                                         & 4,259,780                                    & 8                                                           & 17                                                          \\ 
Palestine                       & ps                             & 2,854                                                      & 6,004,791                                    & 4,820,335                                         & 4,263,491                                    & 2                                                           & 12                                                          \\ 
Qatar                           & qa                             & 5,047                                                      & 11,824,490                                   & 7,891,425                                         & 6,867,304                                    & 2                                                           & 2                                                           \\ 
Somalia                         & so                             & 78                                                         & 168,136                                      & 131,944                                           & 104,946                                      & 8                                                           & 9                                                           \\
Sudan                           & sd                             & 1,162                                                      & 2,348,325                                    & 1,522,274                                         & 1,171,866                                    & 14                                                          & 27                                                          \\ 
Syria                           & sy                             & 1,630                                                      & 2,992,106                                    & 2,184,715                                         & 1,889,455                                    & 12                                                          & 19                                                          \\ 
Tunisia                         & tn                             & 227                                                        & 460,268                                      & 362,806                                           & 239,769                                      & 10                                                          & 10                                                          \\ 
UAE                             & ae                             & 14,923                                                     & 36,121,319                                   & 23,309,788                                        & 18,484,296                                   & 7                                                           & 15                                                          \\ 
Yemen                           & ye                             & 2,391                                                      & 4,783,144                                    & 3,368,262                                         & 3,013,517                                    & 8                                                           & 8                                                           \\ \hline
\multicolumn{2}{|c|}{\textbf{Total}}     & \textbf{233,105}                   & \textbf{507,318,355} & \textbf{318,174,122}      & \textbf{277,430,807} & \textbf{235}                        & \textbf{664}                        \\ \hline
\end{tabular}
\center 
\caption{\label{font-table} Statistics of our data representing 233,105 users from 664 cities and 21 countries. We process more than half a billion tweets, from a larger pool of $\sim$ 6 billion tweets, to acquire our final dataset. Note that the number of states and cities is further reduced after our manual user verification. Eventually, we acquire data for 319 cities, belonging to 192. The data represent all 21 Arab countries.}
\label{tab:data_stats}
\end{table*}

\begin{table}[h]
\centering
\begin{tabular}{lccl}
\hline
\textbf{Country} & \textbf{\%vld\_cntry} & \textbf{\%vld\_city} & \textbf{\#tweets} \\ \hline
\textbf{Algeria} & 77.49\ & 69.74\ & 185,854 \\
\textbf{Bahrain} & 83.95\ & 39.51\ & 25,495 \\
\textbf{Djibouti} & 68.42\ & 68.42\ & 3,939 \\
\textbf{Egypt} & 92.66\ & 64.02\ & 463,695 \\
\textbf{Iraq} & 51.50\ & 37.61\ & 59,287 \\
\textbf{Jordan} & 83.61\ & 54.10\ & 17,958 \\
\textbf{KSA} & 96.37\ & 62.88\ & 353,057 \\
\textbf{Kuwait} & 84.30\ & 34.88\ & 65,036 \\
\textbf{Lebanon} & 92.42\ & 56.06\ & 37,273 \\
\textbf{Libya} & 75.48\ & 72.03\ & 128,152 \\
\textbf{Maurit.} & 45.00\ & 35.00\ & 3,244 \\
\textbf{Morocco} & 75.59\ & 62.42\ & 140,341 \\
\textbf{Oman} & 90.25\ & 77.97\ & 108,846 \\
\textbf{Palestine} & 87.50\ & 82.35\ & 87,446 \\
\textbf{Qatar} & 85.00\ & 77.50\ & 29,445 \\
\textbf{Somalia} & 52.73\ & 45.45\ & 9,640 \\
\textbf{Sudan} & 56.88\ & 41.28\ & 23,642 \\
\textbf{Syria} & 76.28\ & 71.63\ & 79,649 \\
\textbf{Tunisia} & 78.95\ & 75.94\ & 26,300 \\
\textbf{UAE} & 85.31\ & 82.49\ & 129,264 \\
\textbf{Yemen} & 72.41\ & 56.32\ & 47,450 \\ \hline
\textbf{Avg/Total} & 81.00\ & 62.29\ & 2,025,013 \\\hline
\end{tabular}
\caption{Our gold data, from manually verified users.}
\label{tab:gold_data}
\end{table}

\begin{table}[h]
\centering

\begin{tabular}{llll} \hline
\textbf{Country} & \textbf{TRAIN} & \textbf{DEV} & \textbf{TEST} \\ \hline
\textbf{Algeria} & 100,000 & 18,700 & 18,572 \\
\textbf{Bahrain} & 20,387 & 2,556 & 2,552 \\
\textbf{Djibouti} & 3,158 & 408 & 373 \\
\textbf{Egypt} & 100,000 & 46,136 & 46,325 \\
\textbf{Iraq} & 47,395 & 5,903 & 5,989 \\
\textbf{Jordan} & 14,413 & 1,826 & 1,719 \\
\textbf{KSA} & 100,000 & 35,312 & 35,106 \\
\textbf{Kuwait} & 52,127 & 6,416 & 6,493 \\
\textbf{Lebanon} & 29,821 & 3,641 & 3,811 \\
\textbf{Libya} & 100,000 & 12,847 & 12,803 \\
\textbf{Maurit.} & 2,579 & 338 & 327 \\
\textbf{Morocco} & 100,000 & 14,118 & 13,862 \\
\textbf{Oman} & 87,048 & 10,807 & 10,991 \\
\textbf{Palestine} & 69,834 & 8,668 & 8,944 \\
\textbf{Qatar} & 23,624 & 2,968 & 2,853 \\
\textbf{Somalia} & 7,678 & 1,023 & 939 \\
\textbf{Sudan} & 18,929 & 2,334 & 2,379 \\
\textbf{Syria} & 63,668 & 7,987 & 7,994 \\
\textbf{Tunisia} & 21,164 & 2,599 & 2,537 \\
\textbf{UAE} & 100,000 & 13,089 & 12,768 \\
\textbf{Yemen} & 37,886 & 4,833 & 4,731 \\\hline
\textbf{Total} & 1,099,711 & 202,509 & 202,068 \\ \hline
\end{tabular}%
\caption{Distribution of classes in our data splits.}
\label{tab:splits}
\end{table}

\begin{table*}[]
\centering
\begin{tabular}{|l|ll|ll|ll|}
\hline 
\textbf{Setting} & \multicolumn{2}{c|}{\textbf{City}} & \multicolumn{2}{c|}{\textbf{State}} & \multicolumn{2}{c|}{\textbf{Country}} \\ \hline   
\textbf{Eval Metric} & \textbf{acc} & \textbf{F1} & \textbf{acc} & \textbf{F1} & \textbf{acc} & \textbf{F1} \\ \hline 
\textbf{Baseline I (majority in TRAIN)} & 1.313 & 0.008 & 3.110 &	0.032 &  9.191	& 0.802  \\ \hline 
\textbf{Baseline II (single task Attn-BiGRU)} & 2.110	& 0.450   & 3.840 &	0.360 & 21.250 &	7.390  \\ \hline
\textbf{MTL (common-attn)} & 4.070\ & 1.714\ & 5.634\ & 2.152\ & 28.297\ & 13.404\ \\
\textbf{MTL (spec-attn)} & 4.083\ & 1.593\ & 5.921\ & 2.196\ & 29.082\ & 14.058\ \\
\textbf{HA-MTL (city first)} & \underline{12.384}\ & \underline{11.791}\ & \underline{13.696}\ & \underline{12.894}\ & 40.784\ & 30.090\ \\
\textbf{HA-MTL (country first)} & 11.214\ & 10.289\ & 13.460\ & 12.696\ & \underline{40.942}\ & \underline{30.179}\ \\ \hline 
\textbf{BERT} & \textbf{19.528}	& \textbf{19.818}   & \textbf{21.199}\ & \textbf{21.671}\ & \textbf{47.567}\ & \textbf{38.297}\\ \hline

\end{tabular}%
\caption{Performance on DEV. Highest results for MTL are \underline{underlined}. BERT results (best) are in \textbf{bold}.}
\label{tab:sprvsd_eval_dev}
\end{table*}

\begin{table*}[h]
\centering
\begin{tabular}{|l|ll|ll|ll|}
\hline
\textbf{Setting} & \multicolumn{2}{c|}{\textbf{City}} & \multicolumn{2}{c|}{\textbf{State}} & \multicolumn{2}{c|}{\textbf{Country}} \\ \hline
\textbf{Eval Metric} & \textbf{acc} & \textbf{F1} & \textbf{acc} & \textbf{F1} & \textbf{acc} & \textbf{F1} \\ \hline
\textbf{Baseline I (majority in TRAIN)} & 1.313 & 0.008 & 3.110 & 0.032 & 9.191 & 0.802 \\
\textbf{Baseline II (single task Attn-BiGRU)} & 2.740 & 0.880 & 4.450 & 0.910 & 27.170 & 12.820 \\ \hline
\textbf{MTL (common-attn)} & 4.036 & 1.693 & 5.693 & 2.195 & 28.255 & 13.362 \\
\textbf{MTL (spec-attn)} & 4.000 & 1.479 & 5.956 & 2.085 & 28.946 & 13.858 \\
\textbf{HA-MTL (city first)} & \underline{12.295} & \underline{11.736} & \underline{13.728} & 12.836 & 40.349 & \underline{29.869} \\
\textbf{HA-MTL (country first)} & 11.265 & 10.588 & 13.577 & \underline{12.869} & \underline{41.250} & 29.763 \\ \hline
\textbf{BERT} & \textbf{19.329} & \textbf{19.452} & \textbf{19.329} & \textbf{19.452} & \textbf{47.743} & \textbf{38.122}\\ \hline
\end{tabular}%
\caption{Performance on TEST. Highest results for MTL are \underline{underlined}. BERT results (best) are in \textbf{bold}.}
\label{tab:sprvsd_eval_test}
\label{tab:my-table}
\end{table*}

\begin{table}[h]
\centering
 \footnotesize 
\begin{tabular}{lll} \hline 
\textbf{Supervision} & \textbf{acc} & \textbf{F1} \\ \hline 
\textbf{Baseline (majority in TRAIN) } & 9.207 & 0.843 \\
\textbf{Gold} & 46.808	& 37.863 \\
\textbf{Weak} & 41.195 & 23.560 \\
\textbf{Weak+Gold} & \textbf{49.700}	& \textbf{38.651} \\
\textbf{weak\_\textit{then}\_Gold} & 47.862 & 38.560 \\ \hline
\end{tabular}%
\caption{Results on DEV with models exploiting noisy labels on 20 countries (with Djibouti excluded). For comparison, our gold (BERT trained on human-labeled TRAIN) is re-trained with 20 classes.} 
\label{tab:noisy_res_dev}
\end{table}

\begin{table}[h]
\centering
\footnotesize 
\begin{tabular}{lll} \hline 
\textbf{Supervision} & \textbf{acc} & \textbf{F1} \\ \hline 
\textbf{Baseline (majority in TRAIN) } & 9.207 & 0.843 \\
\textbf{Gold} & 46.844 & 37.643 \\
\textbf{Weak} & 41.166 & 23.697 \\
\textbf{Weak+Gold} & \textbf{49.768} & 38.254 \\
\textbf{Weak\_\textit{then}\_Gold} & 47.862 & \textbf{38.560}\\ \hline
\end{tabular}%
\caption{Results on TEST with models exploiting noisy labels on 20 countries (with Djibouti excluded). For comparison, our gold and small-GRU are re-trained with 20 classes.} 
\label{tab:noisy_res_test}
\end{table}

\begin{table}[h]
\centering
\footnotesize 
\begin{tabular}{l|llll}
\hline
\textbf{\#tweets} & \textbf{acc} & \textbf{thresh} & \textbf{F1} & \textbf{thresh} \\ \hline
\textbf{10} & 45.132 & 0.62 & 39.131 & 0.83 \\
\textbf{25} & 54.872 & 0.70 & 48.211 & 0.72 \\
\textbf{50} & 62.006 & 0.65 & 54.833 & 0.75 \\
\textbf{75} & 64.012 & 0.75 & 56.809 & 0.85 \\
\textbf{100} & 65.171 & 0.95 & \textbf{60.335} & 0.95 \\
\textbf{500} & \textbf{66.787} & 0.57 & 58.661  & 0.67  \\
\hline
\end{tabular}%
\caption{User-level evaluation on external data (crawled from the MADAR user base). Note that we take a \textit{thresholded} majority class of predicted tweets as a user-level tag. For thresholding, we use the per-class softmax value in the model's output layer.}
\label{tab:user_eval}
\end{table}

\end{document}